\documentclass[journal,twoside,web]{ieeecolor}  
\usepackage{jsen}

\usepackage{graphics} 
\usepackage{graphicx}
\usepackage{cite}
\usepackage{multirow}
\usepackage{amsmath,amsfonts}
\usepackage{hyperref} 
\usepackage{booktabs}
\usepackage{balance}

\usepackage{algorithmic}
\usepackage{textcomp}
\usepackage{wrapfig}
\usepackage{xcolor}
\usepackage[draft]{changes} 

\usepackage{eso-pic}

\definecolor{abstractbg}{rgb}{0.89804,0.94510,0.83137}
\begin{document}
\title{Pneumatic-Tomographic Tactile Skin for Multicontact Localization and Force Estimation}

\author{Haofeng Chen, Jiri Kubik, Bedrich Himmel, Matej Hoffmann$^\ast$ and Hyosang Lee$^\ast$   %
\thanks{H. Chen, B. Himmel, and M. Hoffmann are with Department of Cybernetics, Faculty of Electrical Engineering, Czech Technical University in Prague, Prague, Czech Republic. (e-mail: matej.hoffmann@fel.cvut.cz)}%
\thanks{J. Kubik is with Department of Computer Science, Faculty of Electrical Engineering, Czech Technical University in Prague, Prague, Czech Republic.}%
\thanks{H. Lee is with Eindhoven University of Technology (TU/e), Eindhoven, The Netherlands. (e-mail: h.lee1@tue.nl)}
\thanks{This work was co-funded by the European Union under the project Robotics and Advanced Industrial Production (reg. no. CZ.02.01.01/00/22\_008/0004590). J.K. was additionally supported by the Technology Agency of the Czech Republic under project No. TN02000028.}
\thanks{$\ast$Equal senior contribution and shared corresponding authors.}
}

\IEEEtitleabstractindextext{%
\fcolorbox{abstractbg}{abstractbg}{%
\begin{minipage}{\textwidth}%
\begin{wrapfigure}[12]{r}{3in}%
\includegraphics[width=2.9in]{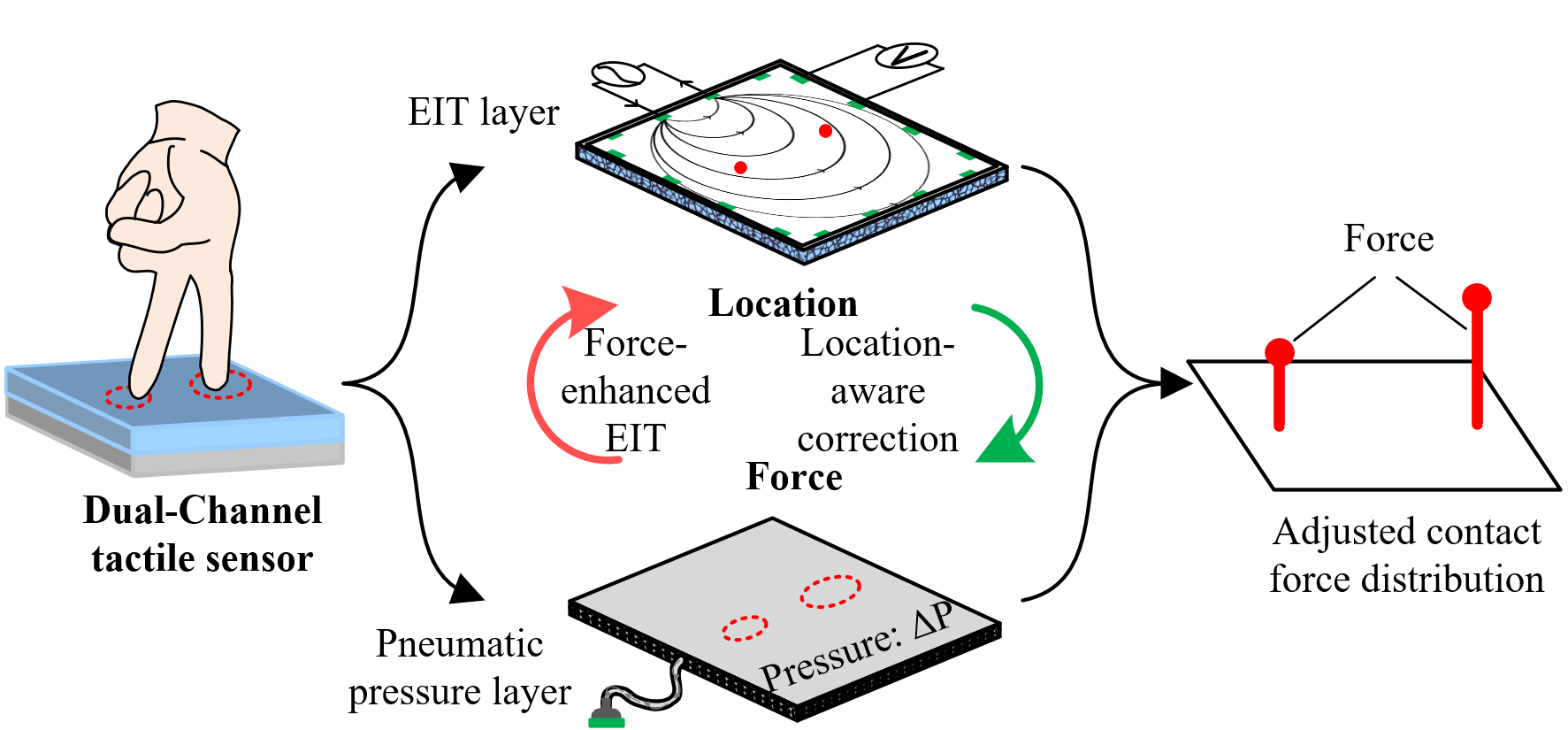}%
\end{wrapfigure}%
\begin{abstract}
Tactile skins based on Electrical Impedance Tomography (EIT) enable large-area contact localization with few electrodes, but suffer from nonuniform sensitivity that limits force estimation accuracy. This work introduces a dual-channel tactile skin that integrates an EIT layer with a pneumatic pressure layer and a calibration framework that leverages their complementary strengths. The EIT layer provides robust multi-contact localization, while the pneumatic pressure layer supplies a stable scalar measurement that serves as contact force estimation. A location-aware correction method is introduced, learning smooth spatial gain and offset fields from a single-session calibration, enabling spatially consistent multi-contact force estimation. With location-aware correction, the proposed system achieves a single-contact force estimation RMSE of 0.59 N across 10–25 mm indenters, representing a 35–60\% reduction over EIT-only approaches (1.45–1.48 N). In multi-contact experiments, the per-contact RMSE is reduced by 39.6\% compared to the uncorrected pneumatic baseline.
The proposed system achieves accurate force estimation across diverse contact configurations, generalizes to varying indenter sizes, and preserves EIT’s inherent advantages in multi-contact localization. By letting the pneumatic pressure layer handle the force estimation and using the EIT layer to determine where each contact occurs, the method avoids the need for large datasets, complicated calibration setups, and heavy machine-learning pipelines often required by previous EIT-only approaches. This dual-channel design provides a practical, scalable, and easy-to-calibrate solution for building large-area robotic skins.
\end{abstract}

\end{minipage}}}

\AddToShipoutPictureFG*{%
  \AtPageUpperLeft{%
    \raisebox{-1.1cm}{
      \makebox[\paperwidth]{%
        \begin{minipage}{0.92\paperwidth}
          \centering
          \color{gray}

          {\fontsize{7}{8}\selectfont
          This is the authors' final version of the manuscript published as:\\
          Chen, H., Kubik, J., Himmel, B., Hoffmann, M. and Lee, H. (2026), 'Pneumatic-Tomographic Tactile Skin for Multicontact Localization and Force Estimation', \\IEEE Sensors 26(16), 23655 - 23665, \textcopyright~IEEE,
          \texttt{https://doi.org/10.1109/JSEN.2026.3704916}\\
         \par}
         \vspace{3pt}\rule{\linewidth}{0.4pt}
        \end{minipage}%
      }%
    }%
  }%
}

\maketitle
\section{INTRODUCTION}
Large-area tactile sensors have great potential for human-robot interaction---in safety as well as social interaction through touch \cite{RN1168,RN444,RN610,RN916,silvera2015artificial}. Seamlessly covering a robot's complex three-dimensional body surfaces is required to perceive contact positions and forces over the complete body. To handle this non-trivial problem, multiple arrays of sensors have been widely adopted \cite{Cheng2019, Cannata2008}. These approaches deploy a large number of sensing modules and dense interconnects, raising manufacturing cost and wiring complexity, possibly increasing susceptibility to electromagnetic interference (EMI) and crosstalk.

\begin{figure}[tp] 
    \centering 
    \includegraphics[width=0.45\textwidth]{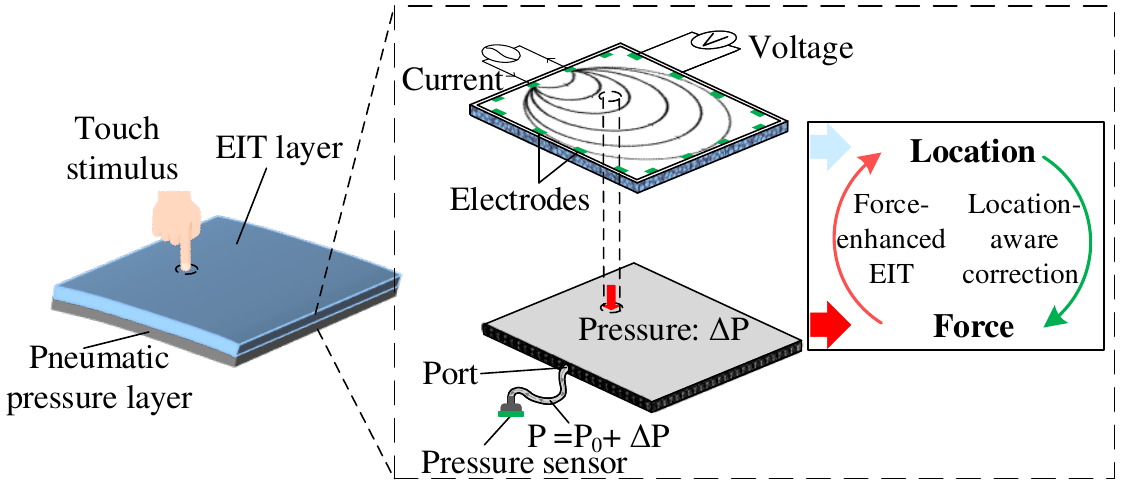} 
    \caption{Overview of the proposed dual-channel tactile skin incorporating an EIT layer and a pneumatic pressure layer.} 
    \label{Fig2} 
\end{figure}

To address these limitations, reconstruction-based methods such as Electrical Impedance Tomography (EIT) have emerged as a promising alternative. EIT-based tactile sensors reconstruct contact-induced conductivity distributions of a large sensing area using a few electrodes. 
This approach offers a cost-effective and feasible solution for large-area tactile sensing \cite{RN1010,RN1020,RN1077,RN543,RN1252}. Despite these advantages, EIT-based tactile sensors have a few challenges arising from the nonuniform sensitivity distribution \cite{RN1068,RN812}, which primarily results from the dependence of the electric field on electrode placement \cite{RN965} and susceptibility to measurement noise and model inaccuracies \cite{RN1086}. 

As the nonuniform sensitivity directly impacts the sensor's contact force estimation, it has been studied by several researchers by using reliable ground-truth contact data to improve physics-based model calibration or to provide training data. Lee et al.~\cite{RN721} introduced a single-point indentation testbed that yields a coefficient field. This coefficient field was then used to calibrate the conductivity image into a pressure field, ensuring approximately uniform sensitivity across the surface, which improves force estimation accuracy. Subsequently, they developed a sophisticated multipoint indenter and sim-to-real transfer learning pipeline for multi-contact prediction \cite{RN1086}. 
Chen and Liu exploited the symmetry of the sensor structure \cite{RN1068}. They proposed the intensity scaling method to correct for the reconstructed conductivity magnitudes, producing a more uniform sensitivity. 
Similarly, Chen et al. \cite{RN929} introduced a pseudoarray method based on single-contact experiment data. This approach equally divided the surface of the EIT-based sensor into a pseudo-array unit, obtained a force-to-conductivity mapping for each unit, and developed the Jacobian vector correction method, which enabled near-uniform sensitivity across the sensing area without extensive calibration data \cite{RN999}. However, their work was limited to the estimation of single-point contact force. 

While these calibration approaches have improved the sensitivity across the surface, making it more uniform, significant challenges remain for EIT-based force estimation. Multi-contact scenarios are especially difficult, often requiring dense spatial calibration and tightly controlled contact conditions to achieve reliable performance.

Integrating additional modalities offers a promising alternative to avoid this calibration challenge. BioTac \cite{RN1253} exemplifies this idea at the fingertip scale: a weakly conductive fluid inside an elastomer deforms under load, producing a distributed impedance pattern, while the pressure of the internal fluid is measured concurrently. This combined impedance and pressure approach is conceptually similar to ours. However, BioTac relies on a sealed fluid capsule with a rigid core and multilayer packaging optimized for fingertips, which complicates scaling to large, conformable skins and does not provide multi-contact force calibration across wide areas. 

Chen et al. \cite{RN1147} introduced a multimodal EIT sensor system with two stacked EIT layers: one layer with a spacer mesh to modulate pressure sensitivity, and another to capture contact geometry. 
While effective, this approach requires two EIT layers, complicating sensor fabrication. 

To simplify the design, pneumatic pressure sensors, such as Park et al. \cite{RN1113}, can be used to provide a better force estimation accuracy than EIT-based tactile sensors.

For industrial application and integration, a pneumatic pressure sensor can provide simple fabrication, reliable force estimation, and safe human-robot collaboration that meet industrial safety certification requirements (see AIRSKIN \cite{zillich2019protection,svarny2022effect}). However, single pressure sensor cannot localize the contact location within the sensor.  

In this work, we therefore introduce a dual-channel tomographic tactile skin that integrates an EIT layer with a thin pneumatic pressure layer in a synergistic dual-channel architecture (Figure~\ref{Fig2}). The EIT layer provides spatial information, such as contact regions and centroids, while the pneumatic pressure layer measures the total contact force, thereby avoiding the inherent nonuniform sensitivity of EIT-based systems. 
Because the pneumatic pressure layer exhibits mild nonuniform sensitivity, we characterize it using single-contact calibration experiments and compute location-aware correction gain and offset fields to correct the mapping between pneumatic pressure and true contact force. 

The key contributions of this work are summarized as follows: 
\begin{itemize}
    \item \emph{Tomographic–pneumatic dual-channel force estimation framework.} We introduce a dual-channel tactile skin and signal processing framework that enhance force estimation by integrating an EIT layer with a pneumatic pressure layer. The complementary modalities provide contact localization and contact force, enabling robust multi-contact force estimation with minimal calibration requirements.
    
    \item \emph{Easy-to-fabricate dual-layer tactile skin.} We design an easy-to-fabricate dual-layer structure in which the EIT and pneumatic layers operate synergistically within a compact, flexible, and large-area form factor, suitable for integration onto robot surfaces.
    
    \item \emph{Location-aware correction fields.} We develop location-aware correction fields that compensates for spatial sensitivity variations within the pneumatic pressure layer and distributes forces across contacts using EIT conductivity change summation, enabling improved quantitative multi-contact force estimation.
\end{itemize}
The rest of this article is organized as follows. Section \ref{sectionII} introduces the dual-channel force calibration framework, including the EIT reconstruction and ROI segmentation, the location-aware correction method for the pneumatic pressure layer, and the hardware fabrication. Section \ref{Parameter characterization} presents the parameter characterization and calibration data collection. Section \ref{sectionIV} reports experimental results for both single- and multi-contact force estimation. Section \ref{sectionV} concludes with a discussion of limitations and future directions.

\section{Force Calibration Framework and Sensor Design}
\label{sectionII}
This section first introduces the overall dual-channel force calibration framework, then details the EIT reconstruction and regions of interest (ROI) segmentation, followed by the force calibration of the pneumatic pressure layer, and finally the dual-layer hardware design. 

\begin{figure*}[htbp] 
\centering 
\includegraphics[width=0.85\textwidth]{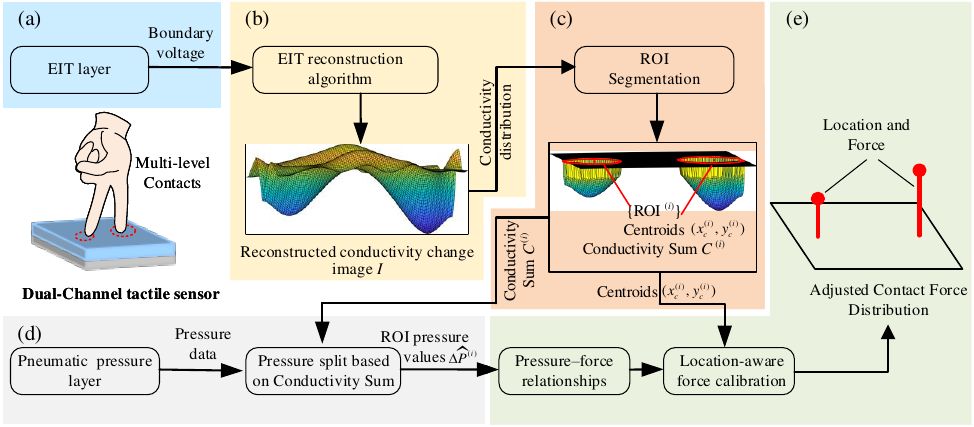} 
\caption{Overview of the proposed dual-channel calibration framework using the EIT layer and pneumatic pressure layer: (a) EIT boundary voltage acquisition, (b) conductivity change image reconstruction, (c) ROI segmentation and centroid extraction, (d) pneumatic pressure measurement and proportional splitting across ROIs, and (e) location-aware force correction yielding the adjusted contact force distribution.} 
\label{Fig:Overview_calibration} 
\end{figure*}

Figure \ref{Fig:Overview_calibration} presents the proposed dual-channel calibration framework for the estimation of contact force. The framework fuses information from the EIT layer and the pneumatic pressure layer in a complementary manner. The EIT layer first acquires boundary voltages (a) and reconstructs a conductivity change image via the EIT reconstruction algorithm (b). The reconstructed image is then segmented into ROIs (c): for each ROI, a binary mask is obtained, its intensity-weighted centroid is extracted, and the conductivity change sum is calculated. In parallel, the pneumatic pressure layer measures the total pressure change (d), which is distributed across the ROIs in proportion to their conductivity sums (for single-contact cases, the entire pneumatic pressure is assigned to the single ROI). Each distributed pressure estimate is then converted to force and refined by the location-aware correction fields (e), yielding the adjusted contact force distribution. In this scheme, the pneumatic pressure layer supplies the force signal while the EIT layer provides contact centroids for spatial correction, enabling accurate per-contact force estimates without compromising EIT's large-area localization capability.

\subsection{Reconstruction principle}
EIT sensing requires solving coupled forward and inverse problems. In the forward problem, the primary objective is to compute the boundary voltage given a set of known parameters, including the internal conductivity distribution, electrode configurations, and applied current patterns at the sensor periphery. We adopt the complete-electrode model to account for contact impedance. Let $\Omega \subset \mathbb{R}^2$ denote a bounded domain representing the sensor area. The sensor includes $D$ silver electrodes ${\{e_l\}}_{l=1}^{D}$ uniformly distributed along the boundary $\partial \Omega$. Let $\sigma(x,y)$ be the conductivity distribution and $\phi(x,y)$ be the electric potential inside $\Omega$. The forward problem can be mathematically formulated as follows:

\begin{equation}
\nabla\cdot[\sigma(x,y)\nabla\phi(x,y)]=0,(x,y)\in\Omega
\label{eq_fwd1}
\end{equation}
with the boundary condition
\begin{equation}
\int_{e_{l}}\sigma(x,y)\frac{\partial\phi(x,y)}{\partial \nu}dS=I_{l},l=1,\ldots,D
\label{eq_fwd2}
\end{equation}

\begin{equation}
\sigma(x,y)\frac{\partial\phi(x,y)}{\partial \nu}=0,(x,y)\in\bigcup_{l=1}^{D}e_{l}
\label{eq_fwd3}
\end{equation}

\begin{equation}
\sigma(x,y)\frac{\partial\phi(x,y)}{\partial \nu}=0,(x,y)\in\partial\Omega\setminus\bigcup_{l=1}^{D}e_{l}
\label{eq_fwd4}
\end{equation}

\begin{equation}
\phi(x,y)+\sigma(x,y)z_{l}(x,y)\frac{\partial\phi(x,y)}{\partial \nu}=\mathrm{~constant~}=V_{l}
\label{eq_fwd5}
\end{equation}
where $\nu$ represents the outward unit normal vector, $I_l$ is the injected current on the electrode $e_l$, $V_l$ represents the measured boundary voltage, and $z_l$ denotes the corresponding contact impedance. To ensure a unique and valid solution to the system, two fundamental conditions must be satisfied regarding charge conservation and reference voltage selection:
\begin{equation}
\sum_{l=1}^DI_l=0,\sum_{l=1}^DV_l=0
\label{eq_bc1}
\end{equation}

The finite element method (FEM) \cite{RN514} is commonly employed to solve equation \eqref{eq_fwd1}. This numerical approach discretizes the continuous domain into a mesh consisting of $N$ finite elements. We adopt a time-difference imaging approach. The linearized forward model describes the relationship between the conductivity change $\Delta\boldsymbol{\sigma}\in\mathbb{R}^N$ and the induced boundary voltage change  $\Delta \boldsymbol{V}\in\mathbb{R}^M$ between the reference and observation time points \cite{RN656}:
\begin{equation}
\Delta \boldsymbol{V}=\mathbf{J}\Delta\boldsymbol{\sigma}+\mathbf{n}_{noise}
\label{eq_fw2}
\end{equation}
where $\mathbf{n}_{noise}$ denotes the noise matrix. The Jacobian matrix $\mathbf{J}\in\mathbb{R}^{M\times N}(M\ll N)$ is also called the sensitivity matrix.

To obtain the conductivity change, we solve the following regularized inverse problem:
\begin{equation}
\Delta\boldsymbol{\sigma}^\ast
= \arg\min_{\Delta \boldsymbol{\sigma}}
\left\| \Delta \mathbf{V} - \mathbf{J}\,\Delta\boldsymbol{\sigma} \right\|^2
+ \gamma\,\|R(\Delta \boldsymbol{\sigma})\|^2 ,
\label{eq1}
\end{equation}
where the regularization term $R(\Delta \boldsymbol{\sigma})$ stabilizes the solution for this ill-conditioned problem. The conductivity change is then calculated via Tikhonov/Newton’s One-Step Error Reconstructor (NOSER):
\begin{equation}
\Delta \boldsymbol{\sigma}^\ast = ({\mathbf{J}^T}\mathbf{J} + \lambda^2 \mathbf{R})^{-1} {\mathbf{J}^T} \Delta \mathbf{V},
\label{eq2}
\end{equation}
with regularization matrix $\mathbf R$ from the NOSER prior \cite{RN983} and scalar parameter $\gamma$ chosen heuristically. This formulation directly solves the inverse problem by mapping voltage differences to conductivity changes.

\subsection{ROI segmentation}
The reconstructed conductivity change $\Delta \boldsymbol{\sigma}^\ast$ is defined on the FEM mesh. To enable image-based processing, it is converted into a regular 2D image:
\begin{equation}
\mathcal{I} = \mathcal{P}(\Delta \boldsymbol{\sigma}^\ast),
\end{equation}
where $\mathcal{P}(\cdot)$ denotes the FEM-to-image mapping, which is implemented in the open-source EIT toolbox EIDORS \cite{RN514}.

The image is normalized to $[0,1]$, yielding $\mathcal{I}_{\text{norm}}$. Contacts are segmented using Otsu’s method \cite{otsu1975threshold}, which yields a threshold $t^\star$. We obtain an initial binary mask

\begin{equation}
\mathcal{B}(x,y)=
\begin{cases}
1,  \mathrm{if} \ \mathcal{I}_{\mathrm{norm}}(x,y)<t^{\star} \\
2,  \mathrm{if} \ \mathcal{I}_{\mathrm{norm}}(x,y) \geq t^{\star}.
\end{cases}
\label{eq_r_b}
\end{equation}

To refine the masks and remove noise, we apply morphological opening and closing \cite{gonzales1987digital}. Let $A$ be a binary image and $B_s$ a disk-shaped structuring element, chosen as a simple and isotropic operator for removing small artifacts and smoothing contact regions. Its radius is chosen to be 2\% of the image size. The operations are defined as
\begin{equation}
\begin{aligned}
A\circ B_s = (A \ominus B_s) \oplus B_s\\
A\bullet B_s = (A \oplus B_s) \ominus B_s
\end{aligned}
\label{eq_mo}
\end{equation}
where $\circ$ means opening operation and $\bullet$ means closing operation, $\ominus$ and $\oplus$ denote erosion and dilation, respectively. 

In our pipeline we apply opening followed by closing to the binary mask image $\mathcal{B}$:
\begin{equation}
M(x,y) = (\mathcal{B} \circ B_s)\bullet B_s .
\label{eq_mask}
\end{equation}

For multi-contact cases, the connected components of $M(x,y)$ are
identified, and the binary mask corresponding to the $i$-th ROI is
denoted by $M^{(i)}(x,y)$.  
The segmented conductivity image for ROI~$i$ is therefore
\begin{equation}
\mathcal{I}^{(i)}(x,y) = \mathcal{I}(x,y)  M^{(i)}(x,y) .
\label{eq_sum}
\end{equation}

The contact location associated with ROI~$i$ is computed as the
intensity-weighted centroid of $\mathcal{I}^{(i)}(x,y)$:
\begin{equation}
(x_c^{(i)},  y_c^{(i)}) =
\left(
\frac{\sum_{x,y} \mathcal{I}^{(i)}(x,y)  x}{\sum_{x,y} \mathcal{I}^{(i)}(x,y)},\;
\frac{\sum_{x,y} \mathcal{I}^{(i)}(x,y)  y}{\sum_{x,y} \mathcal{I}^{(i)}(x,y)}
\right),
\label{center}
\end{equation}
where the summation runs over all pixels belonging to ROI~$i$.

\subsection{Location-aware correction for pneumatic force estimation}
\label{sec:parameter-characterization}

The reconstructed conductivity change is not a reliable proxy for force due to nonuniform sensitivity and boundary effects.  
To obtain an absolute force measurement, we use the pneumatic pressure layer, which provides a pneumatic pressure change relative to a no-load reference state defined as $\Delta P = P - P_0$, where $P$ is the current measured internal pressure and $P_0$ is the initial internal pressure.
 
The applied normal force $F$ can be expressed directly as a function of $\Delta P$. Under quasi-static indentation, $\Delta P$ increases monotonically with the applied load, enabling us to identify an empirical mapping $F = f(\Delta P)$, which we fit using a quadratic function:

\begin{equation}
f(\Delta P)=a_2(\Delta P)^2+a_1 \Delta P+a_0,
\label{eq:F0}
\end{equation}
where $a_0, \ a_1,$ and $a_2$ are coefficients obtained by least-squares fitting on the calibration data.

Although the pneumatic pressure layer yields a single pressure measurement, its mechanical response is not perfectly uniform across the surface due to edge stiffness and shell geometry. The EIT layer, however, provides accurate estimates of the contact location, enabling us to characterize these spatial variations. Using the EIT-derived contact coordinates, we collect calibration data at uniformly sampled locations and fit smooth spatial correction fields for gain and offset. For each indentation, we record the contact location $(x_t,y_t)$, the true applied force $F_t$, and the corresponding pneumatic pressure change $P_t$. 

Given the pneumatic pressure–force mapping $f(\Delta P)$ from (\ref{eq:F0}), we model these spatial variations using continuous quadratic fields.  Specifically, $g(x,y)$ represents a location-aware correction gain and $b(x,y)$ a location-aware correction offset. Each field 
is parameterized by quadratic basis functions:
\begin{equation}
g(x,y)=\Phi(x,y)\boldsymbol{\alpha},\qquad
b(x,y)=\Phi(x,y)\boldsymbol{\beta}.
\label{eq_g_b}
\end{equation}
where $\boldsymbol{\alpha}$ and $\boldsymbol{\beta}$ are the corresponding parameter vectors and $\Phi(x,y)=[ 1,\;x,\;y,\;x^2,\;xy,\;y^2 ]$.

Let $(x_t,y_t, \Delta P_t,F_t)$ be the $t$-th sample, $T$ be the total number of samples. We define:

\begin{align}
\mathbf{A} &=
\begin{bmatrix}
f(\Delta P_1) \Phi(x_1,y_1) & \Phi(x_1,y_1)\\
\vdots & \vdots\\
f(\Delta P_T) \Phi(x_T,y_T) & \Phi(x_T,y_T)
\end{bmatrix}, \label{eq:A}\\[4pt]
\boldsymbol{\theta} &=
\begin{bmatrix}\boldsymbol{\alpha}\\ \boldsymbol{\beta}\end{bmatrix}, \qquad
\mathbf{F} =
\begin{bmatrix}F_1\\ \vdots\\ F_T\end{bmatrix}. \label{eq:thetaF}
\end{align}
The parameters
$\boldsymbol{\theta}$ are obtained by ridge-regularized least squares:
\begin{equation}
\arg\min_{\boldsymbol{\theta}}
\|\mathbf{A}\boldsymbol{\theta}-\mathbf{F}\|_2^2
+\lambda_g\|\boldsymbol{\alpha}-\boldsymbol{\alpha}_0\|_2^2
+\lambda_b\|\boldsymbol{\beta}\|_2^2,
\end{equation}
with prior $\boldsymbol{\alpha}_0=[1,0,0,0,0,0]^\top$.  The resulting
$g(x,y)$ and $b(x,y)$ provide smooth spatial correction fields applicable at any
location on the surface. The ridge weights $(\lambda_g,\lambda_b)$ were selected by inner cross-validation on the calibration data.

When multiple contacts occur simultaneously, the reconstructed segmentation image contains $k$ disjoint ROIs. Let $C^{(i)}$ denote the sum of the conductivity change in the \(i\)-th ROI:
\begin{equation}
C^{(i)} = \sum_{(x,y)\in \mathrm{ROI}^{(i)}} \mathcal{I}^{(i)}(x,y).
\label{eq:cond_sum}
\end{equation}
If not otherwise specified, the EIT conductivity change sum mentioned below refers to the summation of the conductivity change within the detected contact ROI.

The pneumatic pressure layer provides a total pressure change $\Delta P$, which is then distributed to each ROI proportionally to its conductivity sum:
\begin{equation}
 {\Delta\widehat  P^{(i)}} = \frac{C^{(i)}}{\sum_{j=1}^k C^{(j)}} \Delta P.
\label{eq:multi_dP}
\end{equation}

Each local pneumatic pressure estimate $\Delta\widehat  P^{(i)}$ is then converted to force  using the location-aware correction fields derived earlier:
\begin{equation}
\widehat F^{(i)} = g( {x_c^{(i)}}, {y_c^{(i)}}) f(\Delta\widehat  P^{(i)})+b( {x_c^{(i)}}, {y_c^{(i)}}),
\label{eq:multi_pred}
\end{equation}
where $(x_c^{(i)},y_c^{(i)})$ is the EIT-estimated centroid of the $i$-th 
contact region, and $f(\cdot)$, $g(\cdot)$, and $b(\cdot)$ are the pneumatic pressure–force mapping, location-aware correction gain and offset fields introduced above.

This decomposition assumes a proportional relationship between the local conductivity change and the contact-induced pressure change within each ROI. 
While this assumption may not hold perfectly in cases of strong cross-talk or nonlinear interactions between neighboring contacts, it provides a physically reasonable first-order approximation. 
The empirical results in Section~\ref{sec:multi_enhance} demonstrate that in our case, this assumption holds. 

\subsection{Sensing system and material fabrication}
\begin{figure}[htbp] 
\centering 
\includegraphics[width=0.40\textwidth]{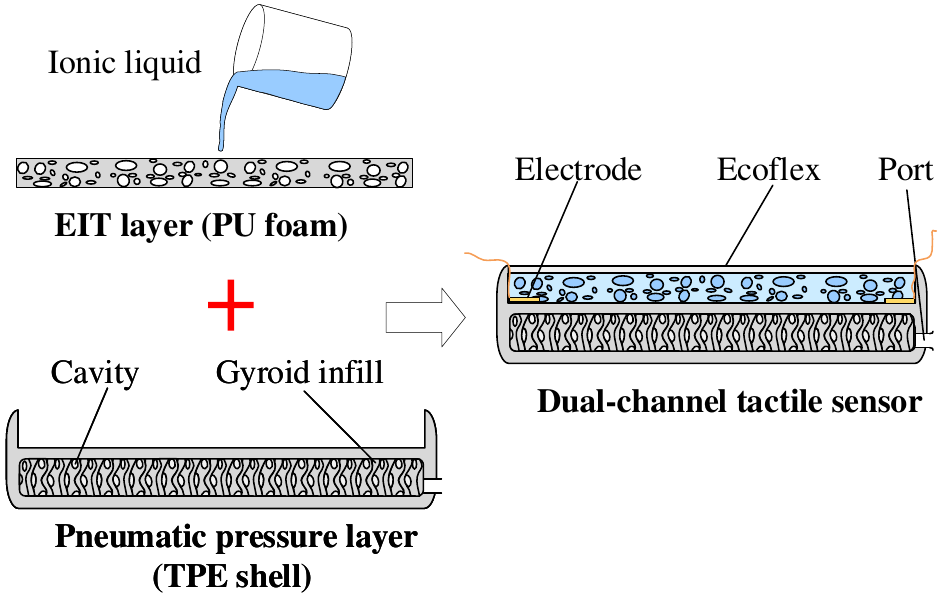} 
\caption{Fabrication process of the flexible EIT-based and pneumatic pressure-based dual-channel tactile sensor.} 
\label{Fig2_2} 
\end{figure}

The EIT layer follows the design of Chen et al. \cite{RN906}, and the pneumatic pressure layer follows Park et al. \cite{RN1113}.
Since the details about each sensor's fabrication, electronics, material characteristics, and signal processing can be found in the literature, this section focuses on the integration of the two sensors in hardware design. 
In order to make the dual-channel tactile skin, the EIT layer and pneumatic pressure layer are fabricated separately. 
The EIT layer is then integrated into the pneumatic pressure layer, which has a slot for encapsulation, as illustrated in Figure \ref{Fig2_2}. 

\subsubsection{EIT layer}
The EIT layer is made of a porous structure and ionic liquid \cite{RN906}. 
The porous structure is polyurethane (PU) foam made using the foaming method. 
On the boundary of the PU foam, sixteen electrodes made of silver tape are attached. 
These electrodes are used to inject an alternating current and measure voltages. 
The PU foam has a square shape with a length of 100 $mm$ and a thickness of 5 $mm$, and the electrodes have a length of 5 $mm$, spaced 25 $mm$ apart from one another. 
After the PU foam is made, the ionic liquid is poured into the foam. 
The ionic liquid is saline water with 125-300 $\mu$S/cm conductivity.
To prevent evaporation of the liquid, an additional elastomer (Ecoflex 00-30, Smooth-On) is then placed to seal the PU foam.

A customized EIT acquisition board uses 4 analog multiplexers (ADG706, Analog Devices), an impedance measurement chip (AD5940, Analog Devices), and a microcontroller unit (STM32F103ZET6, STMicroelectronics). The AD5940 handles current injection and voltage measurement. The ADG706 enables sequential electrode addressing. The STM32F103ZET6 controls the multiplexers, coordinates the AD5940, and transfers data to a computer. Data acquisition follows a two-terminal scheme, where current injection and voltage measurement are performed simultaneously at each electrode pair. With 16 electrodes, 120 independent pair measurements are acquired per frame, resulting in a frame rate of approximately 1.3 frames/s.
The details of the circuitry are explained in the literature \cite{RN906}.

\subsubsection{Pneumatic pressure layer}

The pneumatic pressure layer is created using Fused Filament Fabrication with thermoplastic elastomer (TPE) filament (RUBBERJet-TPE 32, Filament PM) with a Shore hardness of 81A. 
The pneumatic pressure layer is 100 $mm$ $\times$ 100 $mm$ $\times$ 5 $mm$, with a shell thickness of 1 $mm$ and an infill gyroid pattern density of 5$\%$. 
A port connects the shell to a pressure sensor (XGZP6847A005KPG, CFSensor) via pneumatic tubing, and the pneumatic pressure signal is acquired at 20 Hz using a microcontroller (Teensy 4.0, PJRC).

The computational pipeline mainly consists of five stages: EIT reconstruction and conductivity change imaging (49.0 ms), ROI segmentation and binary mask extraction (4.0 ms), centroid estimation and conductivity summation (5.0 ms), pneumatic pressure-to-force mapping (0.1 ms), and proportional pressure splitting with location-aware correction (2.8 ms), giving a mean total processing time of 61.0 ms per frame on a regular laptop (Intel Core i7$-$10750H, 16 GB RAM). The overall throughput is limited by the EIT hardware acquisition rate of 1.3 frames/s, rather than the software processing pipeline, confirming that the proposed framework is compatible with real-time operation at the current hardware frame rate. 

\section{Parameter characterization}
\label{Parameter characterization}
In this section, we characterize how the pneumatic pressure responds to the applied force, forming the basis of our force calibration framework. First, we collect a dataset using controlled single-contact indentations. Next, we characterize the pneumatic pressure–force mapping. Finally, we use the characterized relationships as the basis for learning the location-aware correction fields, which compensate for the mild spatial nonuniformity of the pneumatic pressure layer.

\subsection{Calibration data collection and preprocessing}
\begin{figure}[htbp] 
\centering
\includegraphics[width=0.36\textwidth]{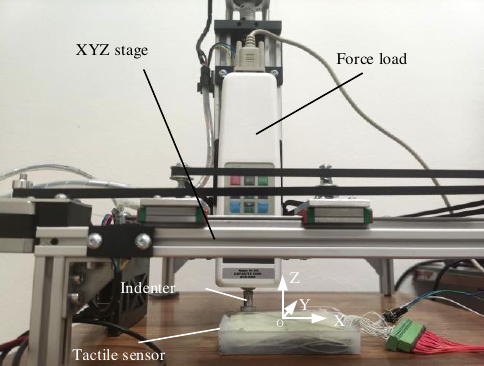} 
\caption{Single-contact experimental setup.} 
\label{Fig:SingleContactSetup} 
\end{figure}

\begin{figure*}[htbp] 
\centering 
\includegraphics[width=0.9\textwidth]{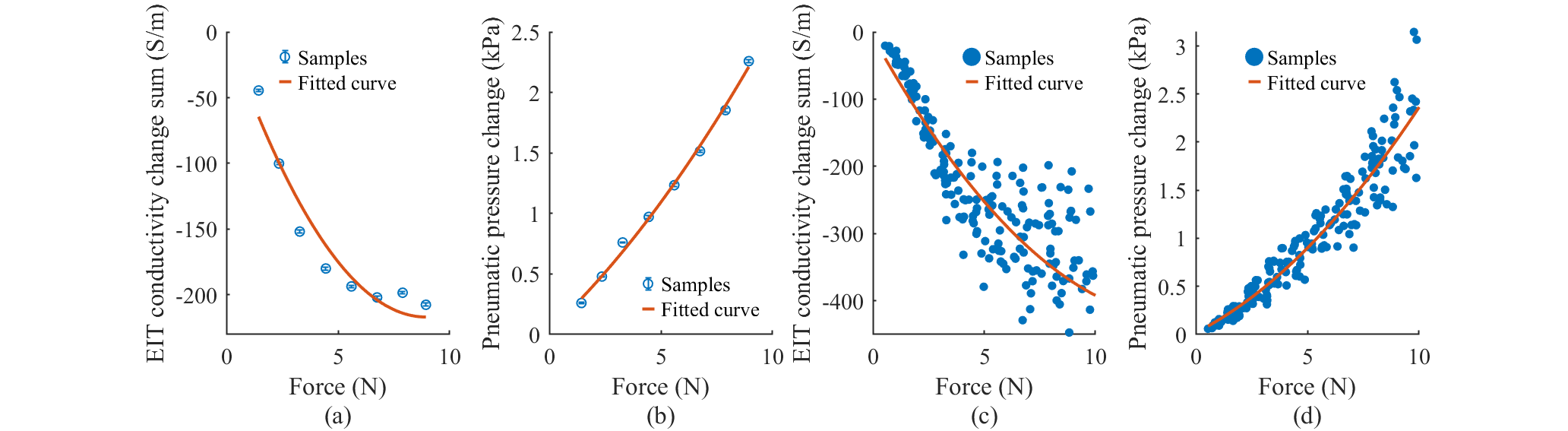} 
\caption{Signal–force relationships between signals and ground-truth force. 
(a) EIT conductivity change sum vs. force at the center location (error bars:  $\pm$1 SD). (b) Pneumatic pressure vs. force at the center location (error bars:  $\pm$1 SD). 
(c) EIT conductivity change sum vs. force using all locations across the sensing area. 
(d) Pneumatic pressure vs. force using all locations. EIT conductivity change sum refers to the summation of the conductivity change within the detected contact ROI.  Under applied force, ionic liquid is expelled from the porous structure in the contact region, which increases the local impedance and results in a negative conductivity change.
} 
\label{fig:fitted_curves} 
\end{figure*}

We first describe how the calibration dataset is acquired and prepared. We used a computer-controlled XYZ stage to apply repeatable indentations onto the sensor surface, as illustrated in Figure \ref{Fig:SingleContactSetup}. A flat-tip cylindrical indenter with a diameter of 15 $mm$ was mounted vertically on the stage to ensure a consistent contact area. The applied normal force was monitored using a load cell (FH100, Sauter), which served as the ground-truth reference. 

To calibrate the relationship between the pneumatic pressure signal and the contact force, we performed indentation trials over a uniform grid of locations on the sensor. A cylindrical indenter (diameter $15$ $mm$) applies normal loads ranging from approximately $1$ N to $10$ N. During each contact, we recorded (i) the ground-truth force from the load cell, (ii) the pneumatic pressure change from the pneumatic pressure layer, and (iii) the raw EIT boundary voltage from the EIT layer. 
At each position and force level, six consecutive frames were acquired within a single indentation and averaged to form each calibration data point. 

In total, 1350 calibration samples are collected. The EIT data were reconstructed to obtain the segmentation image, from which we extracted the contact locations and the conductivity sum within the ROI. Obvious outliers (invalid conductivity peaks or out-of-bounds coordinates) were removed using a simple quality filter, which did not otherwise affect the method.

\subsection{Pneumatic pressure–force relationship}  
\label{Signal–force relationships}

With the calibration dataset prepared, we first characterize how the pneumatic pressure signal relates to the applied force, as this signal serves as the primary basis for force estimation. For comparison, we also plot the EIT conductivity change sum -- force curves to illustrate the contrast between the two signals and to demonstrate why the EIT conductivity change sum is unsuitable for force estimation.
Figure~\ref{fig:fitted_curves} summarizes the relationships between the two signals (changes in pneumatic pressure and EIT conductive sum) and the applied force. Figure~\ref{fig:fitted_curves} (a) and (b) show the signal–force curves when the contact is applied at the center of the sensor, whereas Figure~\ref{fig:fitted_curves} (c) and (d) aggregate data from all contact locations across the surface. 
Each data point in Figure \ref{fig:fitted_curves} (a) and Figure \ref{fig:fitted_curves} (b) represents the mean of six consecutive frames. The mean standard deviation was 0.01 kPa for pneumatic pressure and 0.99 S/m for EIT conductivity sum, confirming good measurement repeatability. Figure \ref{fig:fitted_curves} (c) and Figure \ref{fig:fitted_curves} (d) show the aggregate data across all 25 positions; the spread reflects spatial variability across locations.

At the center location, both signals show a correlation with force, indicating that the EIT conductivity change sum can carry force information locally. However, the EIT conductivity signal to force relationship is nonlinear (Figure~\ref{fig:fitted_curves} (a)); the pneumatic pressure response is nearly linear (Figure \ref{fig:fitted_curves} (b)).

\begin{table}[htbp]
\centering
\caption{Signal–force correlation  ($R^2$)} 
\label{tab:calib_$R^2$}
\begin{tabular}{lcc}
\toprule
Signal & Center location & All locations \\
\midrule
EIT conductivity change sum                  & 0.935      & 0.781 \\
Pneumatic pressure        & \textbf{0.996} & \textbf{0.923} \\
\bottomrule
\end{tabular}
\end{table}

When data from all contact locations across the surface are aggregated, the EIT conductivity change sum-force relationship becomes less pronounced, reflecting the strong nonuniform sensitivity (Figure~\ref{fig:fitted_curves} (c)). In contrast, the pneumatic pressure maintains a high correlation across the entire surface (Figure~\ref{fig:fitted_curves} (d)). The signal–force correlation $R^2$ is summarized in Table \ref{tab:calib_$R^2$}.

\subsection{Location-aware correction fields for pneumatic pressure layer}

\begin{figure}[htbp] 
\centering 
\includegraphics[width=0.48\textwidth]{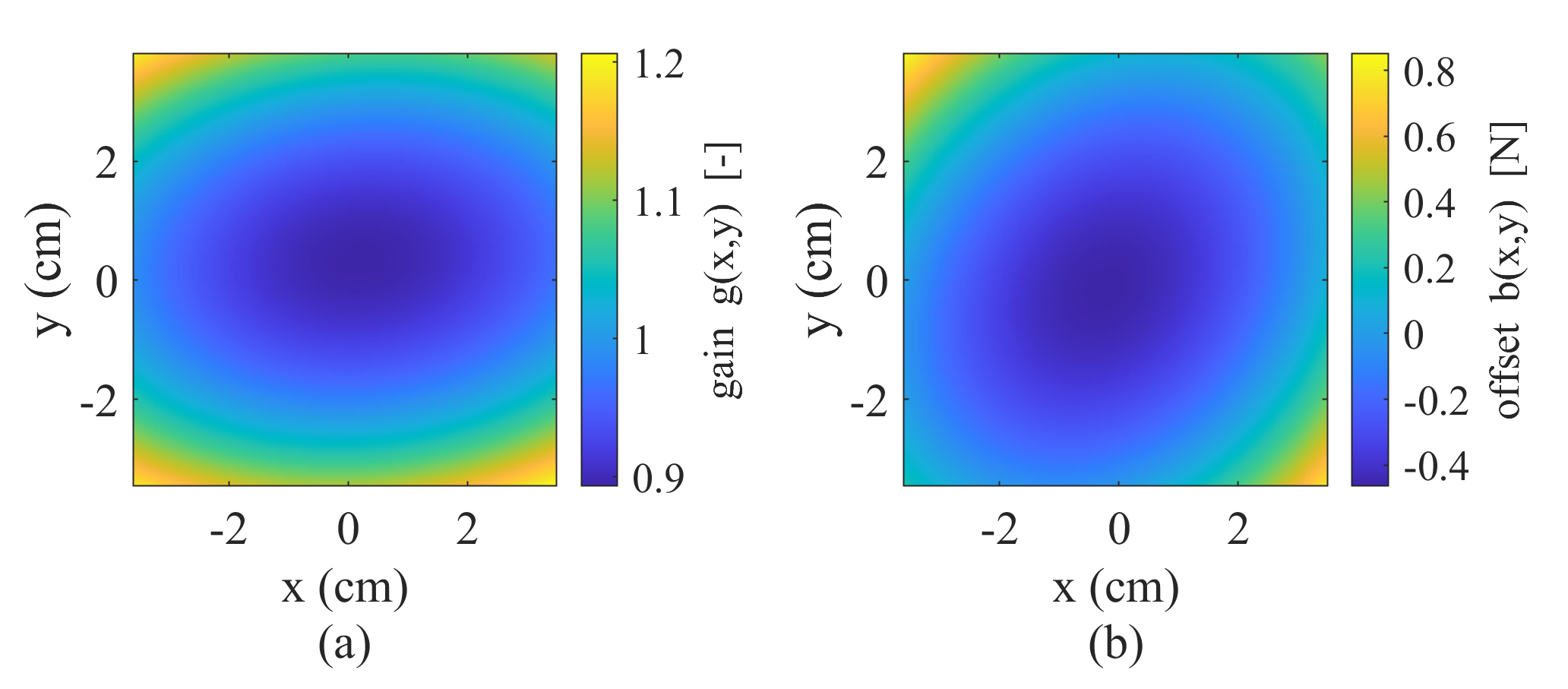} 
\caption{Location-aware correction fields: (a) gain $g(x,y)$ (dimensionless), (b) offset $b(x,y)$ (N).} 
\label{fig:gain_map} 
\end{figure}

We fit location-aware correction fields that modulate the pneumatic pressure–force mapping using the calibration data. In particular, we estimate a location-aware gain field $g(x,y)$ and offset field $b(x,y)$ according to equation\eqref{eq_g_b}. The resulting map (Figure~\ref{fig:gain_map}) reveals the expected nonuniform sensitivity: the pneumatic pressure layer is slightly less sensitive near the boundary due to shell stiffening and geometric constraints, and more sensitive near the center. Accordingly, the learned gain field is higher near the edges and lower near the center. Without this correction, a calibration performed only at the center would systematically under- or over-estimate force depending on contact location.

\section{Results -- Force estimation}
\label{sectionIV}

Our results illustrate the deployment of the complete pipeline (see Figure~\ref{Fig:Overview_calibration} for an overview), with and without the location-aware correction, for a single- and multi-contact case.   

\subsection{Single Contact}
\begin{figure}[htbp] 
\centering 
\includegraphics[width=0.4\textwidth]{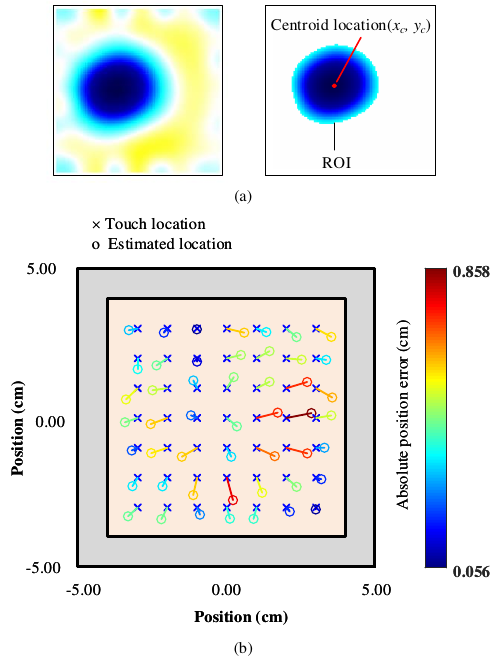} 
\caption{Contact localization error evaluation under single contact. (a) Example of centroid extraction from a reconstructed conductivity image. (b) Position error results over a $7\times7$ grid of test locations, comparing ground-truth indenter positions and estimated contact centroids.} 
\label{PE_single} 
\end{figure}

A constant normal force of 5 N was applied across a grid of $7\times7$ test points on the sensor surface using the XYZ setup illustrated in Figure \ref{Fig:SingleContactSetup}. At each test point, six repeated indentations were performed, yielding 294 samples in total. An example of the centroid extraction procedure is shown in Figure \ref{PE_single} (a).

The position error (PE) was then calculated as the Euclidean distance between the estimated centroid location and the ground-truth indenter position. Figure \ref{PE_single} (b) summarizes the results across the 49 test locations. 
The proposed dual-channel tactile sensor achieved an average position error of 4.4 $\pm$ 1.9 $mm$. This accuracy is comparable to that of EIT-only systems \cite{RN735}.
The spatial resolution of the dual-channel tactile sensor was assessed through the EIT sensing layer using the RES50 and RES75 metrics \cite{tawil2011improveda, RN1010, RN543, park2024graphstructureda}, which quantify the sharpness of the reconstructed contact image relative to the total sensing area. Evaluated across all 49 measurement locations, the proposed sensor achieves RES50 = 69.61 $\pm$ 3.56\% and RES75 = 80.00 $\pm$ 2.92\%.

We now evaluate the force estimation performance of the pneumatic pressure signal, using the EIT conductivity change sum only as a baseline for comparison. In Section~\ref{Parameter characterization}, we characterized the signal–force relationship, and learned location-aware correction gain and offset fields to compensate for spatial nonuniformity. For a fair comparison, we apply the same procedure to both signals: the location-aware correction fields are learned once for each signal type and then held fixed during testing. This allows us to examine two questions: (i) whether pneumatic pressure provides a more reliable force estimation than the EIT conductivity change sum, and (ii) whether the location-aware correction consistently improves force estimation for both layers. 

To evaluate these questions, we compare four force estimation strategies. The evaluation uses independent test sets collected with 10, 20, and 25 $mm$ indenters across multiple loads and locations (529 samples in total).
The cylindrical indenters used in all experiments are flat-tip, fabricated using fused deposition modeling (FDM) with polylactic acid (PLA) material. 
Each strategy corresponds to a different combination of signal type (EIT conductivity change sum or pneumatic pressure) and correction method (with or without the location-aware correction):

(a) \textbf{EIT-NoCal}: 
Force estimated directly from the EIT conductivity change sum using the fitted relationship (Figure \ref{fig:fitted_curves} (c)), without the location-aware correction method.

(b) \textbf{Pneumatic-NoCal}: 
Force estimated from a pneumatic pressure using the fitted relationship (Figure \ref{fig:fitted_curves} (d)), without the location-aware correction method.

(c) \textbf{EIT-LocAware}: 
Force estimated from the EIT conductivity change sum using the fitted relationship (Figure \ref{fig:fitted_curves} (c)), with the location-aware correction method.

(d) \textbf{Pneumatic-LocAware (proposed)}: 
Force estimated from a pneumatic pressure using the fitted relationship (Figure\ref{fig:fitted_curves} (d)), with the location-aware correction method.

\begin{figure*}[htbp] 
\centering 
\includegraphics[width=0.9\textwidth]{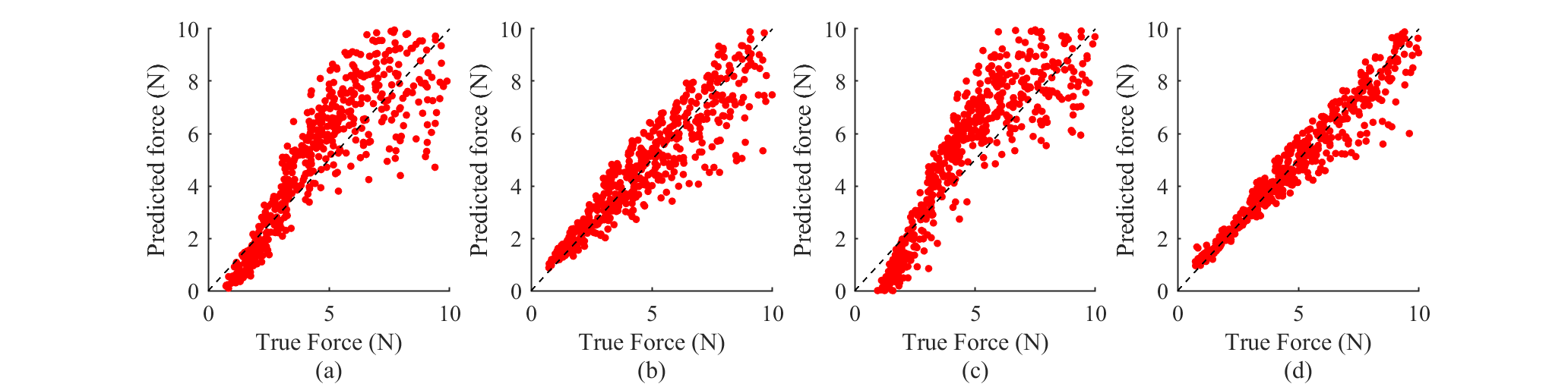} 
\caption{Force estimation on datasets collected with 10 $mm$, 20 $mm$, and 25 $mm$ indenters (red dots; dashed line is the identity). Four strategies are compared: 
(a) EIT-NoCal; 
(b) Pneumatic-NoCal; 
(c) EIT-LocAware; 
(d) Pneumatic-LocAware (proposed). }
\label{fig:predicted_vs_true} 
\end{figure*}

Figure~\ref{fig:predicted_vs_true} compares predicted versus true force for the four strategies. Without location-aware correction (Figure \ref{fig:predicted_vs_true} (a) and (b)), the force estimations deviate from the identity line, but in distinct ways in each strategy. The EIT-NoCal strategy shows large and increasing scatter as the force grows, with predictions spreading widely and becoming less reliable at higher loads. The Pneumatic-NoCal strategy shows a more linear trend, but with moderate scatter across the entire force range.  
After applying location-aware correction (Figure \ref{fig:predicted_vs_true} (c) and (d)), the two strategies behave very differently. For the EIT-LocAware strategy, the correction yields only marginal improvements: the scatter remains large, and low-force predictions (0–3 N) become biased, deviating further from the identity line. This indicates that EIT’s inherent nonuniform sensitivity cannot be adequately corrected through spatial correction. In contrast, the Pneumatic-LocAware strategy shows substantial improvement: the variability is reduced, systematic bias is largely removed, and the predictions cluster closely around the identity line across the entire force range. These results demonstrate that pneumatic pressure offers a far more reliable basis for force estimation and that location-aware correction is highly effective for the pneumatic pressure signal. 

Table~\ref{tab:pred_rmse} reports the quantitative force estimation errors for the four strategies, using the root-mean-square error (RMSE) as the evaluation metric. Consistent with the qualitative trends in Figure~\ref{fig:predicted_vs_true}, the EIT-based strategies yield the highest errors (1.45–1.48 N), with only marginal benefit from location-aware correction. Pneumatic-NoCal achieves a lower error (0.91 N), already surpassing both EIT variants. The proposed Pneumatic-LocAware strategy obtains the best performance, with an RMSE of 0.59 N, representing a 35–60\% reduction in error compared with the other strategies. These results confirm that pneumatic pressure provides a more reliable foundation for force estimation and that location-aware correction is particularly effective for this pathway.  

\begin{table}[t]
\centering
\caption{Cross-diameter force estimation (test on 10/20/25 mm). Lower is better.}
\label{tab:pred_rmse}
\begin{tabular}{lc}
\toprule
Method & RMSE [N] \\
\midrule
EIT-NoCal                & $1.45 \pm 0.89$ \\
Pneumatic-NoCal          & $0.91 \pm 0.62$ \\
EIT-LocAware             & $1.48 \pm 0.83$ \\
\textbf{Pneumatic-LocAware (proposed)} & $\mathbf{0.59 \pm 0.43}$ \\
\bottomrule
\end{tabular}
\end{table}

\begin{figure}[htbp]
  \centering
  \includegraphics[width=0.48\textwidth]{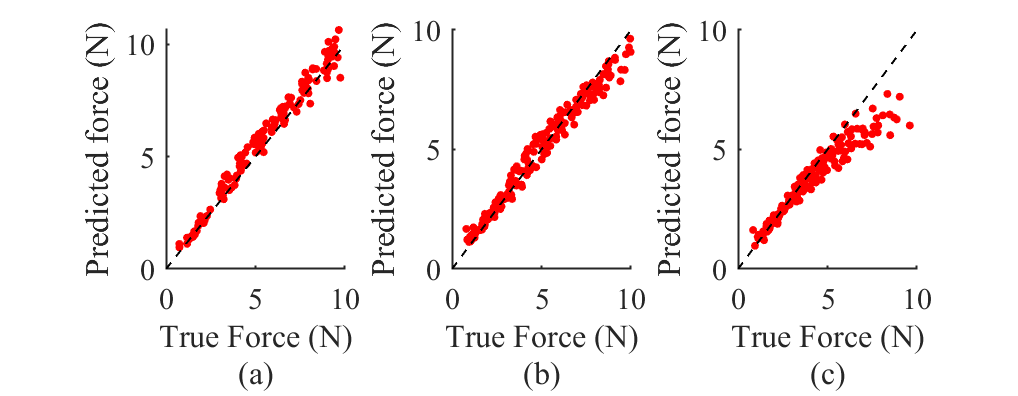}
  \caption{Proposed location-aware correction method evaluated per indenter size.
  Predicted vs.\ true force with identity line (dashed). (a) 10 $mm$;
  (b) 20 $mm$; (c) 25 $mm$.}
  \label{fig:per_diameter_e}
\end{figure}

We further evaluate the Pneumatic-LocAware strategy across different indenter sizes (none of these used during calibration) in Figure~\ref{fig:per_diameter_e}. The predicted forces match the true forces closely for 10 $mm$ and 20 $mm$, with RMSEs of 0.47 N and 0.43 N, respectively. For the 25 $mm$ indenter, the error increases to 0.80 N, mainly at higher loads where the predictions begin to fall slightly below the identity line. This trend is expected: larger indenters distribute the same force over a wider contact area and interact more with the stiffer boundary region, producing a smaller pressure response and thus a modest underestimation of force at higher loads. Despite this, the single calibration trained with a different indenter (15 $mm$) generalizes well. The method remains accurate for 10–20 $mm$ and shows only a modest performance drop for 25 $mm$, consistent with the cross-diameter generalization observed in Figure~\ref{fig:predicted_vs_true}. 
This can be attributed to the force model in Equation \ref{eq:F0} lacking contact-area awareness, compounded by the larger contact area reducing the effective pressure response and the contact region being more likely to interact with the stiffer peripheral zone of the pneumatic shell, which together result in a modest underestimation at higher loads.

\subsection{Multi-contact}
\label{sec:multi_enhance}
\begin{figure}[htbp]
  \centering
  \includegraphics[width=0.3\textwidth]{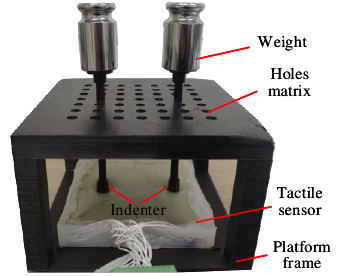}
  \caption{Bench loading platform. A removable weight mounted on a vertical indenter passes through a 7×7 holes matrix to index contact positions, pressing quasi-statically onto the tactile sensor housed in a rigid platform frame.}
  \label{fig:demo_setup}
\end{figure}
Having evaluated single-contact force estimation, we next assess whether the same calibration approach extends to multi-contact conditions. Our goal is to test whether the Pneumatic-LocAware strategy can continue to provide improved force estimates when multiple contacts occur simultaneously. 

Figure~\ref{fig:demo_setup} shows the bench platform. A removable weight (200/300/500g) is mounted on a vertical indenter that passes through a 3D-printed hole matrix (7\(\times\)7 grid, 1.5 cm spacing), allowing repeatable selection of contact locations. The weight applies a quasi-static load through the indenter onto the tactile sensor housed in a rigid platform frame. A fixed cylindrical indenter of 10 $mm$ diameter was used for all contacts to ensure consistency. Using this platform, we applied loads at predefined locations spanning both central and edge regions, with simultaneous multi-contacts.    
We defined eight two- and three-contact configurations (Table \ref{tab:cases}) that span: equal vs. different weights, central vs. near-edge vs. corner contacts, and spread vs. clustered arrangements.

\begin{table}[htbp]
\centering
\caption{Ground-truth configurations for multi-contact experiments.}
\label{tab:cases}
\begin{tabular}{c c c}
\toprule
Case & Contact positions (cm) & Forces (N) \\
\midrule
1 & $(-3,0), (3,0)$ & $0.98, 0.98$ \\
2 & $(-3,-3), (3,-3)$ & $1.96, 1.96$ \\
3 & $(3,3), (-3,-3)$ & $4.91, 4.91$ \\
4 & $(-3,0), (3,0)$ & $1.96, 2.94$ \\
5 & $(3,-3), (0,0)$ & $1.96, 2.94$ \\
6 & $(-3,0), (3,3)$ & $1.96, 4.91$ \\
7 & $(-3,0), (3,-3), (3,3)$ & $1.96, 4.91, 2.94$ \\
8 & $(-3,-3), (3,-3), (3,3)$ & $1.96, 1.96, 2.94$ \\
\bottomrule
\end{tabular}
\end{table}


For each case, the indenters were simultaneously applied to the predefined positions, and six continuous samples were collected during the contact.
Ground-truth force was set by calibrated masses converted to Newtons. 

We evaluate multi-contact force estimation using two strategies from the single-contact study. The EIT-based strategies are not considered here because they exhibited substantially higher errors in the single-contact experiments (Table~\ref{tab:pred_rmse}) and did not provide reliable force estimates.  The remaining strategies are: Pneumatic-NoCal and Pneumatic-LocAware (proposed).
\begin{figure*}[htbp] 
\centering 
\includegraphics[width=0.90\textwidth]{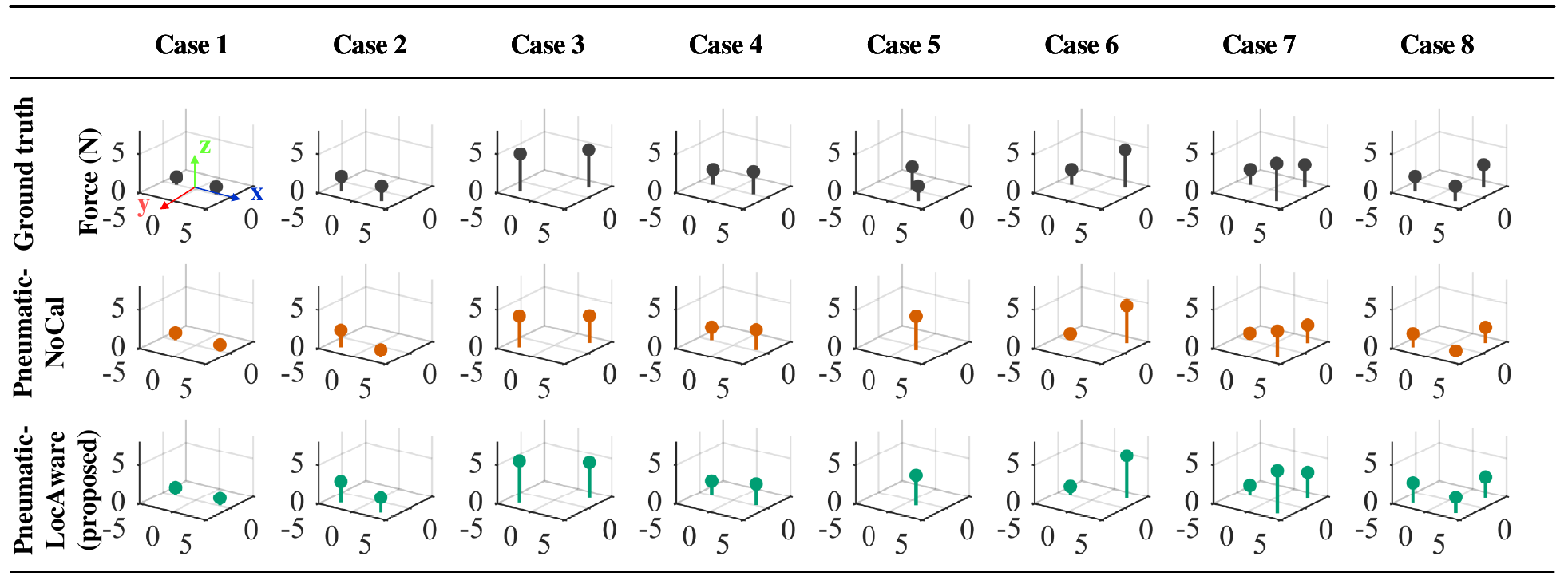} 
\caption{Visualization comparison of multi-contact force calibration across eight representative cases.} 
\label{fig:multicase_vis} 
\end{figure*}
Figure \ref{fig:multicase_vis} illustrates eight representative multi-contact cases. For each case, the top row shows the ground-truth contact locations and forces, the middle row shows the predicted forces from Pneumatic-NoCal, and the bottom row shows the predicted forces from Pneumatic-LocAware (proposed). 

Both strategies can estimate the rough locations and magnitudes of the forces. However, clear differences appear when contacts are close together or near the boundaries. The Pneumatic-NoCal strategy often assigns incorrect force levels to individual contacts because it lacks any spatial correction. In contrast, the Pneumatic-LocAware strategy provides force estimates that are much closer to the ground truth, preserving both the overall magnitude and the relative contribution of each contact.  Case~5 represents the most challenging configuration: the two contacts are close enough that they partially overlap, making the per-contact separation less distinct. This degradation is expected because overlapping contacts reduce the distinguishability of the pressure response. Even in this condition, Pneumatic-LocAware remains more accurate than Pneumatic-NoCal. 

\begin{figure}[htbp] 
\centering 
\includegraphics[width=0.45\textwidth]{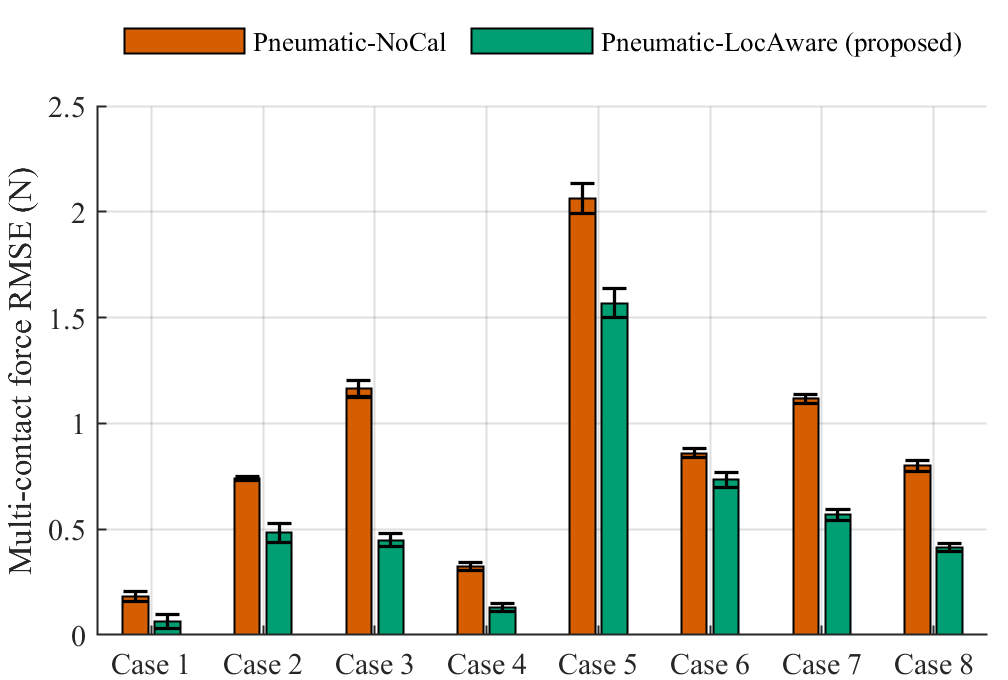} 
\caption{Quantitative comparison of multi-contact force RMSE across the eight multi-contact cases.} 
\label{fig:multicase_bar} 
\end{figure}

Across all resolvable multi-contact trials, the proposed sensor showed an average error of 5.8 mm in contact position (which is about 5.8\% of the sensor width), computed as the average per-trial sum of per-contact centroid errors. This reflects the spatial accuracy under realistic multi-contact conditions where multiple touches occur simultaneously. The quantitative force results are summarized in Figure~\ref {fig:multicase_bar}. The per-contact force RMSE for multi-contact was defined as:

\begin{equation}
\mathrm{per-contact\ RMSE} = \sqrt{\frac{1}{K}\sum_{i=1}^{K} \big( F^{(i)} - \hat{F}^{(i)} \big)^2},
\end{equation}
where $F^{(i)}$ and $\hat{F}^{(i)}$ denote the ground-truth and estimated force of the $i$-th contact, and $K$ is the number of contacts. 

Across the eight multi-contact cases, the $\mathrm{per-contact\ RMSE}$ decreased from $0.91 \pm 0.58$ N with the Pneumatic-NoCal strategy, and to $0.55 \pm 0.47$ N with the Pneumatic-LocAware strategy, corresponding to relative improvements of 39.6\%. Across all resolvable multi-contact trials, the force estimation errors (computed as the average relative difference between the estimated and ground-truth forces) decreased from 26.64\% under the Pneumatic-NoCal to 15.32\% with the proposed Pneumatic-LocAware method. This demonstrates that incorporating location-aware correction substantially improves multi-contact force estimation and remains effective across diverse spatial configurations.
However, a dispersion is observed in the per-contact RMSE across cases. The dispersion in per-contact RMSE across cases reflects the inherent nonuniform sensitivity of EIT-based sensing \cite{RN1068,RN812,RN1086}, where the same force produces different conductivity responses at different locations. This introduces case-dependent errors in the proportional pneumatic pressure-splitting method (Eq. \ref{eq:multi_dP}), and is further compounded when contacts are closely spaced (e.g., Case 5), reducing ROI separability. The dual-channel design mitigates this by offloading force estimation to the pneumatic layer, though residual dispersion remains under challenging configurations.

\section{Discussion and Conclusion}
\label{sectionV}

This paper presents a dual-channel tactile skin that integrates an EIT layer with a pneumatic pressure layer for accurate multi-contact force estimation. By assigning complementary roles to each modality—contact localization to EIT and force measurement to pneumatics—the system overcomes the fundamental limitations of relying on either modality alone. Location-aware correction fields, obtained from a single calibration session, compensate for spatial sensitivity variations and enable consistent per-contact force estimates across the entire sensing surface.

Experiments demonstrate that this tomographic–pneumatic force estimation framework yields accurate force estimation across diverse contact configurations and generalizes to unseen indenter sizes. With location-aware correction, the proposed system achieves an RMSE of 0.59 N on 10–25 mm indenters, improving over the uncorrected pneumatic baseline (0.91 N) and substantially outperforming EIT-only approaches (1.45–1.48 N). In multi-contact tests, the per-contact RMSE decreases from 0.91 N to 0.55 N (a 39.6\% reduction), while the EIT layer maintains robust localization accuracy ($4.4 \pm 1.9$ mm) and clean segmentation.

Compared with state-of-the-art EIT-only approaches \cite{RN721,RN1086,RN1068,RN929,RN999}, which often require large-scale datasets, multi-indenter calibration setup, or complex machine-learning pipelines, our method achieves competitive or superior performance using a simple hardware configuration.  
This highlights the efficiency and practicality of using a pneumatic pressure layer for force estimation instead of relying solely on EIT. In contrast to BioTac \cite{RN1253} or multimodal tomography approaches \cite{RN1147,RN1066}, we rely on a single, low-cost pneumatic pressure layer for force estimation, keeping the hardware simple and practical to build while remaining scalable to large areas.
Furthermore, a pneumatic pressure layer has the potential to be certified similarly to AIRSKIN \cite{zillich2019protection}, rendering a robot covered with such a skin safe for human-robot collaboration.

Despite these advantages, several limitations remain. The proportional pneumatic pressure-splitting rule assumes that the EIT conductivity change sum correlates linearly with local load, which may become inaccurate under large or overlapping contacts. This assumption can also contribute to case-dependent errors when the EIT sensitivity distribution varies across different contact configurations. The pneumatic pressure layer exhibits mild hysteresis, and boundary stiffening, only partly corrected by the learned gain field. Furthermore, because the system does not estimate contact area explicitly at runtime, force accuracy depends to some extent on the indenter geometry used during calibration. Finally, mechanical nonuniformities and mounting constraints can introduce nonuniform biases not captured in the correction model.

Future work will incorporate contact-area awareness into the force model, so that the mapping f($\Delta P$) can adapt to varying contact footprints and further reduce force estimation error for large-diameter indenters. In addition, non-linear pressure-splitting strategies will be investigated to relax the proportionality assumption underlying Eq. \ref{eq:multi_dP}, and improved ROI separation methods will be developed to better handle closely spaced contacts. A systematic multi-device reliability study will also be conducted to characterize device-to-device variability and further validate the robustness of the fabrication process. These developments aim to further enhance the accuracy and robustness of dual-channel tactile skins for real-world robotic applications. 
%





\bibliographystyle{IEEEtran}
\bibliography{Bibliography/reference}

\end{document}